\documentclass{article}

\usepackage{microtype}
\usepackage{graphicx}
\usepackage{subfigure}
\usepackage{booktabs} 

\usepackage{hyperref}

\usepackage[accepted]{icml2025}

\usepackage{amsmath}
\usepackage{amssymb}
\usepackage{mathtools}
\usepackage{amsthm}

\usepackage[capitalize,noabbrev]{cleveref}

\theoremstyle{plain}

\theoremstyle{definition}

\theoremstyle{remark}

\usepackage[textsize=tiny]{todonotes}

\icmltitlerunning{Submission and Formatting Instructions for ICML 2025}

\begin{document}

\twocolumn[
\icmltitle{SciFaultyQA: Benchmarking LLMs on Faulty Science Question Detection with a GAN-Inspired Approach to Synthetic Dataset Generation}


\icmlsetsymbol{equal}{*}

\begin{icmlauthorlist}
\icmlauthor{Debarshi Kundu}{equal,yyy}
\end{icmlauthorlist}

\icmlaffiliation{yyy}{Department of CSE, Pennsylvnia State University, State College, USA}

\icmlcorrespondingauthor{Debarshi Kundu}{dqk5620@psu.edu}

\icmlkeywords{Machine Learning, LLM, Synthetic Dataset, Benchmark}

\vskip 0.3in
]



\printAffiliationsAndNotice{Work in progress}
\begin{abstract}
Consider the problem: ``If one man and one woman can produce one child in one year, how many children will be produced by one woman and three men in 0.5 years?" Current large language models (LLMs) such as GPT-4o, GPT-o1-preview, and Gemini Flash frequently answer "0.5," which does not make sense. While these models sometimes acknowledge the unrealistic nature of the question, in many cases (8 out of 10 trials), they provide the nonsensical answer of "0.5 child." Additionally, temporal variation has been observed: if an LLM answers correctly once (by recognizing the faulty nature of the question), subsequent responses are more likely to also reflect this understanding. However, this is inconsistent.

These types of questions have motivated us to develop a dataset of science questions, SciFaultyQA, where the questions themselves are intentionally faulty. The dataset contains both text based questions and image and text combined questions. We observed that LLMs often proceed to answer these flawed questions without recognizing their inherent issues, producing results that are logically or scientifically invalid. By analyzing such patterns, we developed a novel method for generating synthetic datasets to evaluate and benchmark the performance of various LLMs in identifying these flawed questions. We have also developed novel approaches to reduce the errors.

\end{abstract}

\section{Introduction}
\label{Introduction}
We asked a simple question, when we ask a wrong question even by mistake which doesn't make any sense why do LLMs answer it, doesn't it cost unnecessary cmputation power waste in turn energy waste - carbon footprint increase. Does LLMs understand the faulty questions or just blindly starts solving. Shouldn't it ask before solving ? 

To find this answer we started playing with multiple such nonsesical questions and then math problems and then finally science questions. We thought of building a benchmark for this kind of faulty questions to evaluate whether LLMs can understand questions are itslef wrong or not before going to solve them. 

While generating such faulty questions, we realized generating such data manually is huge time consuming and creates a bias in type of questions. Therefore, we started investigating how LLMs can be used to generate such dataset ? However, if you just ask a LLM to generate  faulty questions and then ask whether the generated question is faulty or not (in a new chat), most of the time it will catch that the question is faulty. This is obvious. Because LLM is creating faulty questions based on its understaning level of which is faulty and which is not. Thus a faulty question generated by one LLM has level of faultiness same as which it can detect. 

But, if we take two different LLMs, ask one LLM say LLM1 to generate faulty question and then ask another LLM, say LLM2 to answer it, a lot of time the LLM2 can not detect the faulty question and gives answer to the faulty question.  

This also shows that different LLMs are good in different fields. Therefore, if one LLM  can 

Recently AI models have been doing well in Science QA. Such as GPQA\cite{rein_gpqa_2023}, ScienceQA\cite{lu2022learn}, SciQ\cite{welbl_crowdsourcing_2017}, ARC-Challenge\cite{clark_think_2018}, SciQA\cite{auer_sciqa_2023} datasets, etc.

Recently GPT-o1 model has broken \cite{openai_learning_2024} human-level  performance in the GPQA dataset \cite{rein_gpqa_2023}. There are multiple other cases where AI models are surpassing the human benchmark\cite{maslej2024aiindex}. This requires the development of advanced benchmarks that can test advanced AI models  \cite{jones_ai_2024}. So far humans have been the gold standard. But in the future AI models will surpass human-level performances  \cite{jones_ai_2024}, \cite{maslej2024aiindex}. How will we create benchmarks for future AI models? This brings us to develop a novel synthetic data generation procedure where AI models will generate new data by competing against each other. We propose a GAN-style synthetic data generation methodology.  

\subsection{Motivation for novel methodology for benchmark creation: Running out of Benchmarks}
We point out a critical problem of running out of good benchmarks. 
The AI models are getting better and soon they will be better than human. Researcher have been developing new tougher benchmark questions as the ML models getting better and better. and human standered has been the golden standered. But what will happen when ML models will surpass those benchmarks. We need to create methodologies to generate new and new benchmarks using SOTA ML models. A new model will be SOTA only when it beats the benchmark generated by previous ML models. 
\subsection{Key Contributions}

\begin{enumerate}
    \item \textbf{Novel Dataset Creation Methods:}
    We curated a new dataset of faulty questions to assess LLMs' ability to recognize and respond appropriately to flawed questions. Alongside this, two novel techniques for generating synthetic datasets were developed.

    \textbf{GAN-inspired dataset generation:} \\
    \textbf{Diffusion-inspired dataset generation: WIP} 

    We want to inject faults into the valid questions using diffusion process. It will be like stylistic \cite{lyu_fine-grained_2023} and compositional change \cite{lyu_styleptb_2021} of sentences. While designing the experiments we asked the question does LLM really understand which part of the question is responsible for logically correctness, which part is responsible for making the question numerically correct or any other type of correctness (we listed few type of faults : TBD)
    \textbf{Can we finetune a model to generate faulty questions?}
We explore if we can fine-tune a LLM using the previously generates synthetic data.
If we generate a new synthetic dataset and then fine-tune or train a new model and then generate another synthetic data, and train another LLM, if we keep iterating this process will the final LLM be capturing a completely different distribution? Like having a diffusion kind of approach.
    \textbf{What makes a question faulty} : We ask the question of what makes a question faulty. In how many fundamental ways someone can make a correct question faulty? Is it finite or infinite? Depend on domain of knowledge ? type of questions - subjective or objective or MCQ?

    \item \textbf{LLM Evaluation:} We systematically evaluated the performance of different LLMs, measuring their ability to detect and handle faulty questions. Our findings indicate that current LLMs exhibit varying degrees of expertise across different types of fallacies.
    
    \item \textbf{Proposed Error-Reduction Methods:} To address these challenges, we proposed several strategies:
    \begin{itemize}
        \item \textbf{AI Agents:} Creating multi-agent systems where multiple LLM models verify each other’s responses before delivering a final answer. This approach leverages the strengths of different models to improve overall performance.
        \item \textbf{Tool Integration:} Incorporating external tools such as WolframAlpha, calculators, fact-checkers, and online search engines into chatbots can significantly enhance their ability to identify and respond appropriately to faulty questions.
        \item \textbf{Harnessing Model Specializations:} Our data suggests that different LLMs have varying areas of expertise. By combining multiple models in a multi-agent framework, we can harness these strengths to create a more robust application capable of effectively addressing flawed questions.
    \end{itemize}

    \item We also investigated whether certain strategies during training or fine-tuning could make LLMs better at recognizing and addressing such cases. Potential methods include:
    \begin{enumerate}
        \item \textbf{Exposure to Faulty Questions:} Introducing flawed questions during training to improve the model's ability to identify and respond appropriately.
        \item \textbf{Enhanced Feedback Mechanisms:} Utilizing reinforcement learning with human feedback (RLHF) or synthetic feedback to refine the model's judgment on logically flawed scenarios.
    \end{enumerate}
\end{enumerate}

\section{Methodology}
\begin{figure}
    \centering
    \includegraphics[width=1\linewidth]{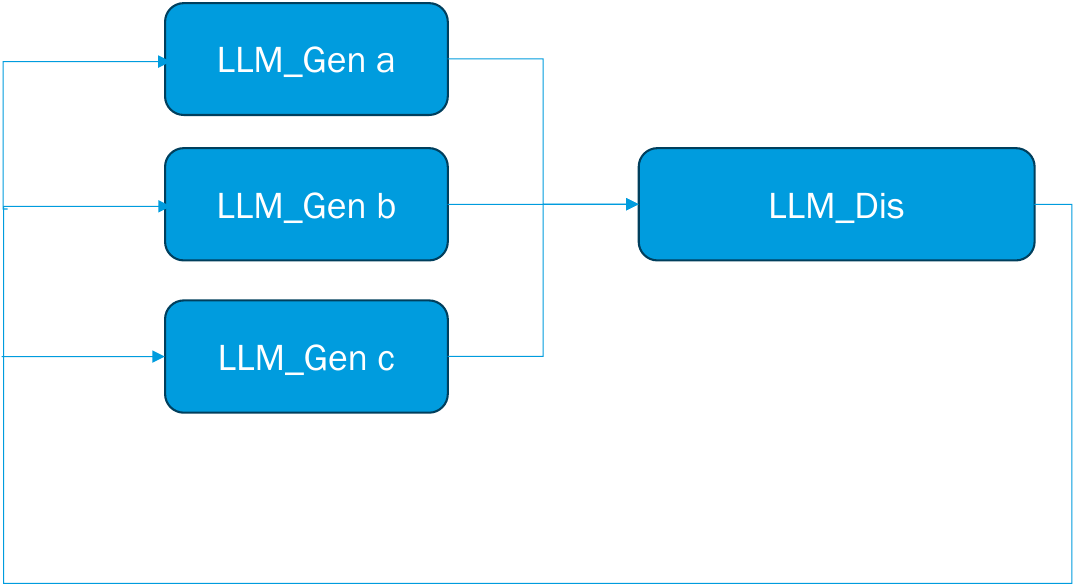}
    \caption{GAN inspired synthetic data generation flow}
    \label{fig:GAN}
\end{figure}

In the \ref{fig:GAN} there are three $LLM\_Gen$ which acts as generators of faulty questions. $LLM\_dis$ stands for discriminator which is used to evaluate if an LLM can detect faulty question or not.

\textbf{GAN-inspired dataset generation:} 
    The technique involves the following steps:
    \begin{enumerate}
        \item Pick a dataset that contains science questions, such as SciQA and SciQ.
        \item Extract each question, its corresponding answer, and any additional available information from the dataset for each row item.
        \item Utilize multiple LLMs, referred to as LLM\_generators (e.g., LLM\_generator1 = GPT 4-o, LLM\_generator2 = Gemini Pro, LLM\_generator3 = Llama 3.1, LLM\_generator4 = Mixtral), to generate faulty versions of the original questions. Each generator also provides a reason why the generated question is faulty and identifies the type of fault, such as logical fallacies, unrealistic scenarios, or violations of physical laws.
        \item Feed the faulty questions generated in step 3 to another LLM, designated as the LLM\_discriminator (e.g., GPT-4). The LLM\_discriminator is not provided with the reasons for faultiness. Instead, it is tasked with analyzing each faulty question one by one to determine if it is indeed faulty and, if so, to explain why. If the question is not deemed faulty, the LLM\_discriminator answers it. These responses (questions and explanations) are then fed back to the LLM\_generators, which use this feedback to refine and generate a new set of faulty questions.
        \item Repeat the process: Each LLM\_generator creates one new faulty question per iteration, following the same steps as before, while the LLM\_discriminator evaluates these new questions. This iterative process continues until the LLM\_discriminator can no longer find faults in any of the generated faulty versions of the original question, or until a predefined maximum number of iterations is reached.
    \end{enumerate}

    Using this iterative approach, we aim to generate faulty versions of the SciQ and SciQA datasets separately. The resulting datasets will include the following columns:
    \begin{itemize}
        \item Science Discipline (and optionally subcategories).
        \item Original Question.
        \item Generated Faulty Question.
        \item type of fault injection (optional)
        \item The reason why the question is faulty.
        \item Faulty Answer by the LLM\_discriminator (applicable only when the discriminator fails to recognize the question as faulty).
    \end{itemize}

After the faulty dataset generated by this method, What if the generated results are not actually faulty but LLM says it is faulty? We need to analyze False Positive cases. 
For this, we took two step process
1. Check with an AI agent which has access to web search
2. Human evaluation 
\section{Experiments and Results}
We evaluated the performance of three LLMs on the generated final dataset SciFaultyQA (1333 rows) 
Here table \ref{tab:llm-evaluation} shows the performance of three different LLMs performance on the generated faulty dataset, SciFaultyQA. 
\begin{table}[h!]
\centering
\begin{tabular}{|l|c|}
\hline
\textbf{Model}       & \textbf{Detection Rate (\%)} \\ \hline
Gemini Flash         & 6\%                         \\ \hline
Llama 3.1            & 12\%                        \\ \hline
GPT 4o               & 16\%                        \\ \hline
\end{tabular}
\caption{LLM Evaluation Results}
\label{tab:llm-evaluation}
\end{table}

Next we tried to improve the performance of chatbots using various techniques. In this case, we only picked the best performing LLM we found in the previous case i.e. GPT 4o. We experimented with three different techniques as mentioned in the table. The corresponding results show that when the LLM has access to the internet it performs much better than without it. We belive it is happening because of access to more information available to it instead of relying on training where it might happen that LLM didn't capture the true concept. But when it accesses the real information, it doesn't make any mistakes.

\begin{table}[h!]
\centering
\resizebox{0.48\textwidth}{!}{%
\begin{tabular}{|l|c|}
\hline
\textbf{Scenario}                          & \textbf{Accuracy (\%)} \\ \hline
Baseline GPT 4o Model Accuracy             & 16\%                   \\ \hline
After Multi-model Agent Implementation     & 30\%                   \\ \hline
With WebSearch Tool Integration            & 65\%                   \\ \hline
\end{tabular}%
}
\caption{Model Accuracy Improvement Across Scenarios}
\label{tab:model-accuracy}
\end{table}

\section{Conclusion}
The \textbf{SciFaultyQA} initiative addresses a critical challenge in the evaluation of large language models (LLMs): their propensity to respond to logically or scientifically flawed questions without recognizing the inherent faults. By developing a novel GAN-inspired approach to generate synthetic datasets, this research introduces a scalable and unbiased method to benchmark LLMs in identifying faulty questions. The results demonstrate varying levels of expertise among LLMs, with significant room for improvement in fault detection capabilities.

Furthermore, the experiments validate the effectiveness of multi-agent systems and tool integrations like web search engines in enhancing LLM performance. For instance, combining models with diverse specializations and providing access to external tools significantly improved accuracy from baseline levels. This highlights the potential of collaborative and augmented frameworks in addressing the limitations of standalone LLMs.

Looking ahead, the research emphasizes the need for further refinement of fault injection techniques, particularly through diffusion-inspired methods, and the exploration of new error-reduction strategies. By continuously iterating on dataset generation and evaluation methodologies, this work lays a foundation for more robust and intelligent AI systems capable of discerning and appropriately handling flawed inputs.
\bibliography{ref_zotero,ref}
\bibliographystyle{icml2025}

\newpage
\appendix
\onecolumn
\section{Appendix:}

GitHub Repository Link : https://github.com/DebarshiKunduPSU/SciFaultyQA


\end{document}